\documentclass{article}

\IfFileExists{neurips_2026.sty}{\usepackage[main,nonanonymous]{neurips_2026}}{}

\usepackage[utf8]{inputenc}
\usepackage[T1]{fontenc}
\usepackage[hypertexnames=false]{hyperref}
\usepackage{url}
\usepackage{booktabs}
\usepackage{amsfonts}
\usepackage{amsmath}
\usepackage{amssymb}
\usepackage{nicefrac}
\usepackage{microtype}
\usepackage{xcolor}
\usepackage{graphicx}
\usepackage{enumitem}
\usepackage{array}
\usepackage{float}

\title{Math Reasoning in LLMs is Organized by Approach,  Not Topic}

\author{Sajjad Goudarzi\\
  Clemson University \\
  \texttt{sgoudar@clemson.edu}
\AND
Samaneh Zamanifard\\
  Clemson University\\
\AND
 Moloud Nasiri\\
  Clemson University \\
\AND
Hamed Rahimian\\
  Clemson University \\
  \texttt{hrahimi@clemson.edu}
}

\begin{document}

\maketitle

\begin{abstract}

Mathematical reasoning benchmarks are typically organized by topic, but language models may organize their internal computation by reusable reasoning approach instead. In this paper, we investigate whether open math-capable LLMs organize internally by topical sub-skill or by reasoning approach, and we present evidence that the approach is the key.

We introduce a {\it generation-replay} protocol: a model first generates a solution, after which we replay the exact prompt-plus-generation trajectory and extract activation-importance signatures over the reasoning tokens. We cluster these signatures without supervision across eight models and five mathematical reasoning sources, then evaluate the recovered structure with structural, semantic, and intervention tests.

Across all 40 model-source cells, the recovered clusters outperform matched-size random baselines. Two independent frontier-LLM judges find approach-level coherence in 77--82\% of real clusters versus 6--11\% in within-source controls, and topic-pure clusters usually receive labels finer than the topic itself. In approach-controlled prompting, changing the requested reasoning approach shifts cluster assignment in seven of eight model conditions, whereas paraphrases largely preserve it.

These results indicate that math-capable LLMs organize internal mathematical computation by reasoning approach rather than benchmark topic. The implication is that topic-stratified benchmarks and topic-balanced training corpora can still miss the axis that matters: even deliberately topic-balanced corpora may remain imbalanced over reasoning approaches.  
\end{abstract}

\section{Introduction}
\label{sec:introduction}

Mathematical reasoning benchmarks usually organize problems by topic or
sub-skill, such as arithmetic, algebra, geometry, probability, or number theory  \citep{saxton2019analysing,cobbe2021training,hendrycks2021measuring,
amini2019mathqa,li2024numinamath}. These labels are useful for dataset construction, sampling, and reporting, but they are external
annotations rather than direct measurements of a model's internal computation.
Prior work also shows that generated rationales are not always faithful descriptions of the computation supporting an answer 
\citep{turpin2023language,lanham2023measuring}. 
This leaves a basic question
unresolved \citep{meng2022locating, burns2022discovering, zou2023representation}: when math-capable LLMs solve a problem, is their internal
computation organized by benchmark topic or by reusable reasoning approaches? We use \emph{reasoning approach} throughout to mean the reusable solution pattern a model executes.

Topics and reasoning approaches need not coincide. 
Problems from different topics can require the same approach, such as
finite-case enumeration, trial-division testing, quantity bookkeeping, or
one-variable isolation. Conversely, a single topic can contain many distinct approaches. If internal mathematical reasoning is primarily topic-organized, then topic-stratified benchmarks and topic-balanced training corpora provide a natural axis for measuring and shaping model competence. If it is instead approach-organized, then topic-level evaluation can obscure the structure that actually governs model behavior: two models with the same topic-level scores may rely on different approach-specific competencies, and a corpus balanced by topic may
remain imbalanced over reasoning approaches. 

We study this question with a \emph{generation-replay} protocol. For each problem, the model first generates its own solution; we then replay the exact prompt-plus-generation trajectory and extract activation-importance signatures over the generated reasoning tokens. 
Because the protocol profiles the model's realized solution trace rather than a reference solution, it targets the computation that the model actually executed. 
We cluster these signatures without using source, topic, sub-skill, correctness, or reference-solution labels. We then test whether the resulting clusters are non-random, semantically approach-coherent, and sensitive to changes in the requested reasoning approach rather than to surface paraphrase.

Across eight open-weight math-capable models and five mathematical reasoning
sources, the evidence converges on approach-level internal organization. In
all 40 model-source cells, the recovered clusters outperform matched-size random baselines. Two independent semantic judges find that real clusters are
far more approach-coherent than within-source controls, and an
approach-controlled intervention shows that changing the requested reasoning
approach shifts cluster assignment in seven of eight model conditions while paraphrases largely preserve it. Together, these results indicate that the
recovered internal axis tracks what the model is doing, not only what the
problem is about.

Our contributions are threefold. First, we formulate and test a concrete
question about internal organization in math-capable LLMs: whether they group
mathematical computation by benchmark topic or by reusable reasoning approach.
Second, we introduce a generation-replay profiling method that clusters
model-generated traces without using source, topic, sub-skill, correctness, or
reference-solution labels. Third, across eight open-weight models and five math reasoning sources, we show that the recovered structure is both non-random and approach-sensitive, implying that topic-stratified benchmarks
and topic-balanced training corpora can miss an important axis of model competence.

\section{Related Work}
\label{sec:related}
Prior work relevant to our question spans three threads: topic-stratified
mathematical reasoning benchmarks, studies of generated reasoning traces, and
representation-level interpretability. Our method sits at their intersection:
we ask whether internal signatures of model-generated math solutions organize
by benchmark topic or by reusable reasoning approach.

Mathematical reasoning is a key testbed for evaluating multi-step symbolic and quantitative computation in LLMs
\citep{li2024evaluating,mirzadeh2024gsm}. Most math benchmarks define competence through external topic labels: arithmetic,
algebra, geometry, probability, and number theory, while the reviewed papers
differ in how tightly these labels constrain reasoning approaches. DeepMind Mathematics provides programmatically generated tasks across
arithmetic, algebra, probability, comparison, and calculus-like categories for
controlled skill-label evaluation \citep{saxton2019analysing}. GSM8K adds
linguistically diverse grade-school word problems requiring short multi-step
arithmetic reasoning and verifier-based solution selection
\citep{cobbe2021training}, while MATH introduces competition-style problems
with full derivations that expose the limits of scale alone
\citep{hendrycks2021measuring}. MathQA and NuminaMath make approaches more
explicit: MathQA links questions to executable operation programs
\citep{amini2019mathqa}, and NuminaMath provides large-scale
competition-level problem--solution pairs with chain-of-thought-style
solutions and domain/question-type metadata \citep{li2024numinamath}. 
Open math-capable models such as DeepSeekMath, Qwen2.5-Math, and DeepSeek-R1
improve verifiable reasoning through math-focused pretraining,
self-improvement, reward modeling, and reinforcement learning
\citep{shao2024deepseekmath,yang2024qwen2,guo2025deepseek}. Yet they are
still evaluated mainly by behavioral metrics---final-answer accuracy, pass
rates, verifier scores, and leaderboards---which show whether a model solves a
problem, not how its internal computation is organized
\citep{cobbe2021training,hendrycks2021measuring,saxton2019analysing,
li2024numinamath}. We instead test whether activation-importance signatures
from model-generated solution traces group by benchmark topic and source or by
reusable reasoning approaches.

Explicit intermediate reasoning can substantially affect model behavior.
Scratchpad training helps Transformers execute multi-step tasks
\citep{nye2021show}, chain-of-thought prompting improves arithmetic and
symbolic reasoning with few-shot demonstrations \citep{wei2022chain}, and
zero-shot chain-of-thought elicits stronger reasoning from a generic
step-by-step cue \citep{kojima2022large}. A central lesson is that the same problem can admit multiple valid reasoning
routes. Self-consistency samples and aggregates multiple paths
\citep{wang2022self}; least-to-most prompting decomposes hard problems into
simpler subproblems \citep{zhou2022least}; tree-of-thought and RAP frame
reasoning as search or planning over intermediate states
\citep{Yaoetal2023,hao2023reasoning}; and program-aided methods express parts
of the solution as executable operations \citep{gao2023pal,chen2022program}.
Together, these methods support an approach-sensitive view: similar solution
approaches can recur across topics, and one problem can be solved by different
approaches. Generated rationales, however, are not necessarily faithful to the computation
that produced the answer \citep{turpin2023language,lanham2023measuring}. This
motivates our generation-replay protocol: rather than infer approach from text
alone, we replay the exact generated tokens and profile activation-importance
signatures along the realized trajectory. Unlike prior uses of reasoning traces
mainly for elicitation, verification, or supervision, we use them as aligned
trajectories for interpretability: if the internal axis is approach-sensitive, then valid changes in the requested reasoning approach for the same problem should shift signatures more than surface
paraphrases do.

A complementary line of work studies model behavior through internal representations rather than generated text alone. 
Mechanistic interpretability localizes computations to transformer components or circuits \citep{meng2022locating,wang2022interpretability,conmy2023towards}, while population-level methods recover broader activation structure through sparse features, representation directions, or latent-knowledge probes \citep{cunningham2023sparse,zou2023representation,burns2022discovering}. Related work also finds interpretable reasoning behaviors in base models when reasoning mechanisms are activated at appropriate times \citep{venhoff2025base}. Our work belongs to this representation-level line rather than circuit identification: given a model's own generated math solution trace, we cluster population-level activation-importance signatures over generated reasoning tokens and test whether the resulting structure is organized by benchmark topic or by reusable reasoning approach. Our signatures are related to attribution methods such as saliency maps, Integrated Gradients, and Grad-CAM \citep{simonyan2013deep,sundararajan2017axiomatic,selvaraju2017grad}; because such methods can yield plausible but unreliable explanations \citep{adebayo2018sanity}, we use activation-times-gradient values only as compact clustering features, not as standalone causal explanations. The evidential burden is instead placed on validation: clusters must outperform matched random partitions, receive approach-specific semantic labels, and respond more to approach-changing prompts than to surface paraphrases. Because approach-level cluster labels are expensive to obtain at scale, C2 uses two frontier-LLM judges; prior work shows that strong language models can approximate human judgments in some settings but can also exhibit position, verbosity, and self-enhancement biases \citep{liu2023g,zheng2023judging}. Our protocol addresses these risks by requiring specific reasoning-approach labels, rejecting generic labels, and evaluating real clusters against matched controls in the same format.

\section{Experimental Setup and Method}
\label{sec:method}
\subsection{Models and corpora}
\label{subsec:models_corpora}

We evaluate eight open-weight math-capable language models selected to vary
both model scale and training regime. The model suite contains three
reasoning-distilled DeepSeek-R1-Distill-Qwen models at 1.5B, 7B, and 14B
parameters; three instruction-tuned Qwen2.5-Instruct models at the same
scales; and two Mistral-family instruction-tuned models,
Mistral-7B-Instruct-v0.3 and Ministral-8B-Instruct-2410. This selection allows
us to test whether the discovered structure is specific to a single model
family or persists across reasoning-distilled and instruction-tuned systems.

The mathematical reasoning sources are chosen to span a
approach-constrained--to--naturalistic axis. At one end, DeepMind
Mathematics contains programmatically generated problems whose source labels often
strongly constrain the required operations. At the other end, competition-style
sources such as MATH and NuminaMath contain broader topics in which several
reasoning approaches may appear under the same label. GSM8K and MathQA occupy
intermediate positions: GSM8K consists of natural-language grade-school
arithmetic problems with relatively narrow quantitative reasoning patterns,
while MathQA includes operation-program annotations that make part of the
reasoning approach explicit. We use the source-provided sub-skill metadata for sampling and validation only; the clustering algorithm never observes these labels.

For each source, we sample up to 200 prompts from each available
source/sub-skill cell. This yields 3,800 prompts per model, as summarized in
Table~\ref{tab:corpora}. The same balanced prompt set is used for all eight
models, producing 30,400 model-problem generations and 40 model-source cells.
Holding the prompt distribution fixed ensures that differences in the resulting
signatures reflect model-specific generated trajectories and internal
activations rather than differences in the sampled problems.

\begin{table}[t]
  \centering
  \small
  \begin{tabular}{llrr}
    \toprule
    Source & Characterization & Sub-skills & Prompts \\
    \midrule
    \texttt{deepmind\_math} &
    Procedurally generated mathematical tasks & 5 & 1{,}000 \\
    \texttt{gsm8k} &
    Grade-school arithmetic word problems & 1 & 200 \\
    \texttt{math\_hendrycks} &
    Competition-style mathematical problems & 6 & 1{,}200 \\
    \texttt{mathqa} &
    Operation-annotated word problems & 2 & 400 \\
    \texttt{numinamath\_1\_5} &
    Competition-level aggregated problem set & 5 & 1{,}000 \\
    \midrule
    Total & & 19 & 3{,}800 \\
    \bottomrule
  \end{tabular}
  \caption{
  Mathematical reasoning sources used for profiling. We sample up to 200
  prompts from each source/sub-skill cell and reuse the same 3,800-prompt
  balanced sample across all eight models.
  }
  \label{tab:corpora}
\end{table}

\subsection{Generation-replay signatures}
For each model--problem pair $(M,x)$, we construct a signature in two
stages. First, the model generates a solution $y=(y_1,\ldots,y_T)$ using
deterministic greedy decoding with a maximum budget of 8,196 generated tokens.
Second, we replay the exact prompt-plus-generation token sequence in a single
teacher-forced forward/backward pass. Replay fixes the realized trajectory,
removing sampling variation from the profiling stage, and ensures that all
features are computed on the same token sequence that the model actually
generated.

Let $R(x,y)$ denote the generated reasoning-token positions. We exclude prompt
tokens and all generated tokens after a fixed answer-boundary detector fires.
The replay loss is the causal language-modeling loss restricted to reasoning
tokens:
\[
\mathcal{L}_{R}(M,x,y)
=
-\frac{1}{|R(x,y)|}
\sum_{t \in R(x,y)}
\log p_M(y_t \mid x, y_{<t}).
\]
Thus, the signature is computed from the model's own solution trace, rather
than from a reference solution or from the final answer alone.
For each recorded module $m$, with reasoning-token activations
$h_{m,t}\in\mathbb{R}^{d_m}$ and parameters $\theta_m$, we compute three
features. The first two summarize activation magnitude over reasoning tokens:
\[
a^{\mathrm{abs}}_m
=
\frac{1}{|R|}
\sum_{t\in R}
\frac{\|h_{m,t}\|_1}{d_m},
\qquad
a^{\mathrm{rms}}_m
=
\left(
\frac{1}{|R|}
\sum_{t\in R}
\frac{\|h_{m,t}\|_2^2}{d_m}
\right)^{1/2}.
\]
The third feature measures activation-times-gradient importance. We first
compute the parameter-normalized gradient norm
\[
g_m
=
\frac{\|\nabla_{\theta_m}\mathcal{L}_{R}\|_2}
{\|\theta_m\|_2+\epsilon},
\]
and then define $
\iota_m = a^{\mathrm{abs}}_m \cdot g_m
$.
The final generation-replay signature is the concatenation
\[
z(x,y,M)
=
\operatorname{concat}_{m}
\left[
a^{\mathrm{abs}}_m,\;
a^{\mathrm{rms}}_m,\;
\iota_m
\right].
\]
Raw gradient norms are not used as standalone features; they enter only through
the activation-times-gradient product. This keeps all structural, semantic, and intervention analyses in the same pre-specified feature space.

\subsection{Unsupervised clustering}
\label{subsec:clustering}

We cluster signatures separately within each model-source cell. For a fixed
cell, let $Z\in\mathbb{R}^{n\times p}$ be the matrix of generation-replay
signatures, with one row per generated solution. Source labels, sub-skill
labels, and correctness labels are not used for clustering.
Each cell is processed with the same unsupervised pipeline:
\[
Z
\;\rightarrow\;
\ell_2(Z)
\;\rightarrow\;
\operatorname{TruncatedSVD}_{20}
\;\rightarrow\;
\ell_2(\cdot)
\;\rightarrow\;
\operatorname{K-Means}.
\]
The number of clusters is selected independently for each cell by maximizing the silhouette score over
$k\in\{5,\ldots,12\}$. K-Means is run with 10 initializations and a fixed random seed. All reported validation tests use the resulting cell-specific partition.

\subsection{Validation protocol}
\label{subsec:validation}

The clustering step produces one unsupervised partition for each model-source
cell. We validate these partitions with three complementary tests. C1 asks
whether the discovered clusters have non-random geometric structure in the
generation-replay feature space. C2 asks whether cluster exemplars share
specific reasoning approaches, using
within-source controls to calibrate the judge. C3 asks whether cluster
assignment tracks the model's chosen reasoning approach by comparing
approach-controlled prompts against paraphrase controls. Together, these tests evaluate whether the clusters exist, whether they admit a coherent approach-level interpretation, and whether they respond to changes in the requested reasoning approach rather than surface wording.

\subsubsection{C1: Matched-size random structural test}
\label{subsec:c1}

C1 tests whether the discovered partitions contain nontrivial geometric
structure in the generation-replay feature space. For each model-source cell,
we compute the silhouette score of the actual K-Means partition in the
SVD-reduced space. We then draw $B=200$ random partitions that preserve the
same cluster-size multiset as the actual partition and compute their silhouette
scores under the same representation.
We use the finite-sample empirical $p$-value:
\[
p_{\mathrm{emp}}
=
\frac{
1+\sum_{b=1}^{B}
\mathbf{1}
\left[
\mathrm{sil}_{\mathrm{rand},b}
\ge
\mathrm{sil}_{\mathrm{actual}}
\right]
}
{B+1}.
\]
A cell passes C1 when $p_{\mathrm{emp}}<0.05$. Preserving the cluster-size
multiset controls for the possibility that the observed silhouette is caused
only by the number or imbalance of clusters rather than by meaningful structure
in the signatures.

\subsubsection{C2: LLM-judge semantic approach test}
\label{subsec:c2}

C2 tests whether the geometrically discovered clusters correspond to coherent
reasoning approaches rather than to arbitrary groups of math solutions. For each cluster, we select the five nearest-centroid exemplars in the SVD-reduced
space. This selection rule is deterministic and avoids hand-picking examples.
Each judge prompt contains the original problem text, the model-generated
reasoning trace, a summary of the cluster composition, and the dominant labels
of sibling clusters.

The judge is asked to return a structured assessment containing a short
candidate approach label, three to five shared features, distinguishing
features relative to nearby clusters, any outlier, an integer count of how
many of the five exemplars follow the named approach, a specificity judgment,
and an integer confidence score. Generic descriptions such as ``algebraic
manipulation'' or ``apply the formula'' are explicitly rejected unless the exemplars share a distinctive sequence of operations.

A cluster is counted as \emph{approach-coherent} for a judge if and only if all three
conditions hold:
\[
\texttt{n\_exemplars\_following\_approach} \ge 4,
\]
\[
\texttt{approach\_label} \neq \texttt{NO\_COHERENT\_CATEGORY},
\]
and
\[
\texttt{specificity}
\in
\{
\texttt{more\_specific},
\texttt{cross\_label}
\}.
\]
A cluster whose label merely recovers the dominant source sub-skill
(\texttt{matches\_label}) is not counted as a C2 success. This rule is
intended to distinguish approach-level coherence from ordinary topic recovery.

To calibrate the judge, we construct within-source controls for each
model-source cell. Each control bundle contains five exemplars drawn
randomly from other clusters in the same model-source cell. Control
bundles are rendered in the same format as real clusters. This tests
whether the judge would assign plausible approach labels to arbitrary
within-source collections of math solutions that do not share a
cluster assignment. The pre-specified C2
success thresholds are a real-cluster coherent rate of at least 60\%, a
control coherent rate of at most 20\%, and a real-control gap of at least
40 percentage points.

\subsubsection{C3: Approach-controlled prompting intervention}
\label{subsec:c3}

C3 tests whether cluster assignment tracks the model's chosen reasoning approach 
rather than the identity or wording of the problem. We construct a separate
mixed-source approach bank from 50 base problems randomly sampled from the
3,800-problem balanced corpus. Each base problem is expanded into one baseline
prompt, two paraphrase controls, and a set of approach prompts that request
different valid reasoning approaches. Across the bank this yields 50 baseline
prompts, 100 paraphrase controls, and 130 approach-prompt variants, for 280
expanded prompts per model.

For each expanded prompt, the model generates a solution under the same
8192-token deterministic decoding budget. We then apply the same
generation-replay profiler to the generated reasoning tokens. Within each
pre-specified model condition, signatures are clustered intrinsically
using the same representation pipeline: 20-dimensional SVD,
$\ell_2$ normalization, and K-Means with silhouette-selected
$k\in\{5,\ldots,12\}$.

Let $c(q)$ denote the cluster assigned to the expanded prompt $q$. For each base
problem $i$, let $b_i$ be the baseline prompt, $\mathcal{P}_i$ its paraphrase
controls, and $\mathcal{M}_i$ its approach-prompt variants. We compute two
metrics.
First, the approach-shift rate measures whether different requested reasoning approaches for the same problem land in different clusters:
\[
\texttt{shift\_rate}
=
\frac{1}{|\mathcal{I}|}
\sum_{i\in\mathcal{I}}
\mathbf{1}
\left[
\left|
\{c(q):q\in\mathcal{M}_i\}
\right|
\ge 2
\right],
\]
where $\mathcal{I}$ is the set of base problems with multiple approach prompts.
Second, paraphrase invariance measures whether
surface rewordings preserve the baseline cluster:
\[
\texttt{paraphrase\_invariance}
=
\frac{
\sum_i \sum_{q\in\mathcal{P}_i}
\mathbf{1}[c(q)=c(b_i)]
}
{
\sum_i |\mathcal{P}_i|
}.
\]

A model condition is confirmed if and only if both pre-specified criteria hold:
$\texttt{shift\_rate}\ge 0.50$, and $
\texttt{paraphrase\_invariance}\ge 0.70$.
A condition is rejected if
$\texttt{shift\_rate}<0.30$ or if paraphrases shift at least as often as
approach prompts; otherwise it is marked partial. The approach-bank conditions are
pre-specified before running the C3 analyzer.

\section{Results}
\label{sec:results}

\subsection{C1: Structural signal in every model-source cell}

C1 tests whether the discovered partitions reflect nontrivial geometry in the
generation-replay feature space, rather than artifacts of the number or size of
clusters. Across all 40 model-source cells, the actual K-Means partition has a
higher silhouette score than every one of the 200 matched-size random
partitions. Thus, the raw exceedance count is $0/200$ in every cell. With the
finite-sample correction defined in Section~\ref{subsec:c1}, this gives the
minimum attainable empirical value,
\[
p_{\mathrm{emp}}=\frac{1}{201}<0.005,
\]
for every model-source cell.

The actual silhouettes are modest but consistently separated from the matched
random baselines. The mean actual silhouette across cells is 0.204, the median
is 0.191, and the range is 0.123--0.323. Table~\ref{tab:c1-source} summarizes
the result by source. DeepMind Mathematics has the highest mean silhouette,
consistent with its tightly templated approach structure, but every source
passes the same structural test.

\begin{table}[!tb]
  \centering
  \small
  \begin{tabular}{lrrrrr}
    \toprule
    Source & Cells & Clusters & Mean $k$ & Mean sil.\ & Sil.\ range \\
    \midrule
    \texttt{deepmind\_math}   & 8 & 54 & 6.75 & 0.283 & 0.241--0.323 \\
    \texttt{gsm8k}            & 8 & 43 & 5.38 & 0.166 & 0.123--0.202 \\
    \texttt{math\_hendrycks}  & 8 & 40 & 5.00 & 0.175 & 0.149--0.252 \\
    \texttt{mathqa}           & 8 & 42 & 5.25 & 0.200 & 0.158--0.249 \\
    \texttt{numinamath\_1\_5} & 8 & 46 & 5.75 & 0.195 & 0.160--0.228 \\
    \midrule
    Total                     & 40 & 225 & 5.62 & 0.204 & 0.123--0.323 \\
    \bottomrule
  \end{tabular}
  \caption{C1 structural result by source. All 40 cells have
  $p_{\mathrm{emp}}<0.005$ against 200 matched-size random partitions.}
  \label{tab:c1-source}
\end{table}

C1 establishes that generation-replay signatures contain stable cluster
structure across models and sources. This test does not by itself identify the
semantic content of the clusters or rule out topic alignment; it shows only
that the partitions are not explained by random assignments with the same
cluster-size profiles. The next analyses ask whether this structure is better
explained by benchmark topics or by reasoning approaches.

\subsection{C2: Clusters reflect reasoning approaches, not topic labels}

C1 establishes that the clusters are real. C2 asks what they
\emph{are}: do they correspond to coherent reasoning approaches?
We first test whether the recovered clusters simply recover
source-provided sub-skill labels. For each cluster $c$, we compute
dominant sub-skill purity
\[
\mathrm{purity}(c)=\max_s \frac{|c\cap S_s|}{|c|},
\]
where $S_s$ denotes a source-provided sub-skill. A cluster is
topic-pure when $\mathrm{purity}(c)>0.70$.

\begin{table}[!tb]
  \centering
  \small
  \begin{tabular}{lrrr}
    \toprule
    Source & Clusters & Topic-pure clusters & Topic-pure rate \\
    \midrule
    \texttt{gsm8k}            & 43 & 43 & 100.0\% \\
    \texttt{deepmind\_math}   & 54 & 28 & 51.9\% \\
    \texttt{mathqa}           & 42 & 11 & 26.2\% \\
    \texttt{numinamath\_1\_5} & 46 &  1 & 2.2\% \\
    \texttt{math\_hendrycks}  & 40 &  0 & 0.0\% \\
    \midrule
    All sources               & 225 & 83 & 36.9\% \\
    Excluding GSM8K           & 182 & 40 & 22.0\% \\
    Broad non-templated sources & 128 & 12 & 9.4\% \\
    Competition-style sources & 86 & 1 & 1.2\% \\
    \bottomrule
  \end{tabular}
  \caption{
  Topic purity of recovered clusters under dominant sub-skill purity
  \(>0.70\). Topic-pure clusters are concentrated in sources where topic
  and approach are expected to be tightly coupled. In broad
  multi-approach sources, recovered clusters almost never simply recover
  benchmark sub-skill labels.
  }
  \label{tab:topic-purity}
\end{table}
Overall, 83/225 clusters (36.9\%) are topic-pure, and this is
concentrated in sources where topic and approach are mechanically
coupled. GSM8K accounts for 43 topic-pure clusters, but it has only
one source-provided sub-skill. DeepMind Math contributes 28/54.
On the broad non-templated sources (MathQA, NuminaMath, Hendrycks
MATH), only 12/128 clusters (9.4\%) are topic-pure; on the two
competition-style sources alone, only 1/86 (1.2\%). The recovered
clusters are therefore not simply recovering benchmark sub-skill
labels. A text-based baseline confirms that this dissociation is not an
artifact of the clustering pipeline: clustering TF-IDF/SVD
representations of the same generated reasoning traces produces
81.8\% topic-pure clusters with a sub-skill ARI of 0.38, compared
with 36.9\% and 0.10 for the activation-based clusters
(Appendix~\ref{sec:text-baseline}). The surface text of the model's
reasoning is organized by topic; the internal activations do not. 

The next question is what they \emph{do} correspond to.
For each model-source cell, the C2 judge evaluates every within-source
cluster produced by the C1 pipeline. Each cluster's five
nearest-centroid exemplars are drawn from the same source; only the control bundles are drawn from other clusters within
the same source rather than from the cluster being evaluated.
Two independent frontier-LLM judges from
different model families (Claude Opus-4.7 and GPT-5.4, both at high
reasoning effort) evaluate all 225 real clusters and 80 within-source
controls under the strict approach-coherence rubric defined in
Section~\ref{subsec:c2}. Neither judge observes the other's output.
Table~\ref{tab:c2-overall} reports the aggregate result.
This semantic evaluation confirms approach-level coherence.
\begin{table}[t]
  \centering
  \small
  \begin{tabular}{lrrrr}
    \toprule
    & \multicolumn{2}{c}{Claude Opus-4.7} & \multicolumn{2}{c}{GPT-5.4} \\
    \cmidrule(lr){2-3} \cmidrule(lr){4-5}
    Bundle type & $n$ & Coherent rate & $n$ & Coherent rate \\
    \midrule
    Real clusters         & 225 & 185/225 = 82.2\% & 225 & 174/225 = 77.3\% \\
    Within-source controls &  80 &   5/80\phantom{0} = 6.2\%  &  80 &   9/80\phantom{0} = 11.2\% \\
    \midrule
    Gap                   &     & 76.0\,pp          &     & 66.1\,pp \\
    \bottomrule
  \end{tabular}
  \caption{Aggregate C2 result under the strict 4-of-5
  approach-count rule. Both judges independently clear all three
  pre-specified thresholds: real coherent rate ${\ge}60\%$, control
  rate ${\le}20\%$, gap ${\ge}40$\,pp.}
  \label{tab:c2-overall}
\end{table}

The approach-organization hypothesis predicts how C2 should vary across
sources. Approach-templated sources, where each sub-skill maps to a
narrow approach set, should show clean within-topic approach
resolution. Broad naturalistic sources, where approaches cross topic
boundaries and clusters mix related variants, should produce lower
coherence under a strict 4-of-5 exemplar rule.
Table~\ref{tab:c2-source} shows that both judges recover this gradient
independently.

\begin{table}[t]
  \centering
  \small
  \begin{tabular}{lrrrrrr}
    \toprule
    & \multicolumn{3}{c}{Claude Opus-4.7}
    & \multicolumn{3}{c}{GPT-5.4} \\
    \cmidrule(lr){2-4} \cmidrule(lr){5-7}
    Source
      & Real & Control & Gap
      & Real & Control & Gap \\
    \midrule
    \texttt{gsm8k}
      & 39/43 = 90.7\% & 1/16 = 6.2\%  & 84.5\,pp
      & 38/43 = 88.4\% & 2/16 = 12.5\% & 75.9\,pp \\
    \texttt{deepmind\_math}
      & 47/54 = 87.0\% & 1/16 = 6.2\%  & 80.8\,pp
      & 44/54 = 81.5\% & 1/16 = 6.2\%  & 75.2\,pp \\
    \texttt{mathqa}
      & 33/42 = 78.6\% & 1/16 = 6.2\%  & 72.3\,pp
      & 31/42 = 73.8\% & 2/16 = 12.5\% & 61.3\,pp \\
    \texttt{numinamath\_1\_5}
      & 35/46 = 76.1\% & 2/16 = 12.5\% & 63.6\,pp
      & 33/46 = 71.7\% & 3/16 = 18.8\% & 53.0\,pp \\
    \texttt{math\_hendrycks}
      & 31/40 = 77.5\% & 0/16 = 0.0\%  & 77.5\,pp
      & 28/40 = 70.0\% & 1/16 = 6.2\%  & 63.8\,pp \\
    \bottomrule
  \end{tabular}
  \caption{C2 by source under both judges. Both judges recover the same
  source-level pattern: approach-templated sources (GSM8K, DeepMind
  Math) produce the highest coherence rates, while broad competition-style
  sources (NuminaMath, Hendrycks MATH) are lower but still well above
  controls. Every source clears the pre-specified thresholds under both
  judges independently.}
  \label{tab:c2-source}
\end{table}

GSM8K clusters expose quantity-bookkeeping approach variants at
approximately 90\% coherence under both judges. DeepMind Math clusters
resolve template-specific approach variants at 81--87\%. The competition-style sources are lower in absolute terms but remain strongly separated from
controls: Hendrycks MATH achieves 70--78\% real coherence against
0--6\% controls, and NuminaMath achieves 72--76\% against 13--19\%
controls.

\subsection{C3: Approach prompts steer the cluster, paraphrases do not}

C1 and C2 establish that the clusters are real and approach-aligned. C3 asks the remaining question: does the
cluster reflect the \emph{problem's identity} or the \emph{model's
chosen reasoning approach}?

\begin{table}[t]
  \centering
  \small
  \begin{tabular}{>{\raggedright\arraybackslash}p{0.50\linewidth}
                  >{\raggedright\arraybackslash}p{0.10\linewidth}
                  >{\raggedright\arraybackslash}p{0.10\linewidth}
                  >{\raggedright\arraybackslash}p{0.10\linewidth}}
    \toprule
    Model condition & C3 verdict & Shift & Paraphrases Invariance \\
    \midrule
    Qwen2.5-1.5B-Instruct &
      confirmed & 0.64 & 0.72\\
    Qwen2.5-7B-Instruct &
      confirmed & 0.64 & 0.84 \\
    Qwen2.5-14B-Instruct &
      confirmed & 0.60 & 0.78 \\
    Ministral-8B-Instruct &
      confirmed & 0.94 & 0.82 \\
    Mistral-7B-Instruct-v0.3 &
      partial & 0.96 & 0.48 \\
    DeepSeek-R1-Distill-Qwen-1.5B &
      confirmed & 0.68 & 0.76 \\
    DeepSeek-R1-Distill-Qwen-7B &
      confirmed & 0.84 & 0.70 \\
    DeepSeek-R1-Distill-Qwen-14B &
      confirmed & 0.78 & 0.72 \\
    \bottomrule
  \end{tabular}
  \caption{C3 approach-bank results. Seven of eight model conditions
are fully confirmed. Mistral-7B-Instruct passes the shift criterion
but fails paraphrase invariance.}
  \label{tab:c3}
\end{table}

The result is positive across model families. Changing the requested
reasoning approach shifts the cluster (shift rates 0.60--0.96 across
all eight conditions). Paraphrases preserve the baseline cluster in
seven of eight conditions (invariance 0.70--0.84). The exception is
Mistral-7B-Instruct, which passes the shift criterion (0.96) but
fails paraphrase invariance (0.48), indicating that its activation
signatures are more sensitive to surface wording.

\section{Discussion}
\label{sec:discussion}
Taken together, C1--C3 support a consistent interpretation: the recovered
generation-replay structure tracks the model's chosen reasoning approach more
than the problem's topic, source, or wording. 
The source-level variation in C2
fits this view. Sources whose labels map tightly to narrow approach families, such as GSM8K
and DeepMind Math, yield the highest coherence rates, whereas
broader competition-style sources remain well above control baselines but score
lower because one topic label can span several related reasoning approaches. 
This extends prior work on mathematical benchmarks, generated reasoning traces, and
representation-level interpretability by identifying the internal axis that
organizes model-generated mathematical reasoning.

The implications are practical. If math reasoning is internally organized by
approach, then topic-stratified accuracy is incomplete: two models with similar topic-level scores may rely on different
mixtures of reasoning approaches, and a topic-balanced training set may still be
imbalanced over the computations that matter. 
Replayed reasoning traces and approach-aware clustering,  therefore, offer aligned trajectories for
evaluation, auditing, intervention, and data curation. 

The present evidence is structural, semantic, and intervention-based, but not
yet circuit-level. A natural next step is to identify components associated
with particular approach clusters and test whether targeted interventions
degrade performance primarily on problems requiring those approaches. A second
direction is practical: testing whether approach-aware data curation or
evaluation improves downstream behavior relative to topic-balanced baselines.
Human audits of cluster exemplars would also strengthen the semantic
interpretation of the discovered approach families.

\section{Limitations}
\label{sec:limitations}

Several limitations remain.
First, the semantic validation in C2 relies on frontier-LLM judges rather than
human ground-truth annotations. Using two independent judges and within-source controls reduces this risk, but does not remove this concern. Second, the intervention
evidence in C3 is prompt-based rather than mechanistic, so it does not identify
the parameters, attention heads, or circuits implementing each approach. Third, the approach-bank experiment is targeted rather than exhaustive, so the results support
approach-sensitivity in aggregate but do not establish uniform effects across all models, sources, or approach families. Fourth, the analysis is limited to
model-generated mathematical reasoning traces from open-weight math-capable models, and may not transfer unchanged to closed models or non-mathematical
domains. Finally, the recovered clusters should be viewed as descriptive signatures of recurring reasoning behavior, not as a unique taxonomy or a complete mechanistic account.

\section{Conclusion}
\label{sec:conclusion}

We tested whether open math-capable LLMs organize their internal
computation by topical sub-skill or by reasoning approach. Across eight models and five sources, three 
validation layers---structural (C1),
semantic (C2), and intervention-based (C3)---point to the same answer. 
Generation-replay signatures contain real structure in all 40
model-source cells (C1); two independent judges find a strong approach-level
coherence in real clusters relative to within-source controls (C2); and
approach-controlled prompting shows that cluster assignment follows the requested reasoning approach more than paraphrase (C3).

These findings suggest that topic-stratified benchmarks, topic-balanced
training corpora, and topic-level evaluation reports do
not fully capture the axis along which models organize mathematical
reasoning. An approach-aware re-stratification, requiring no new
annotations, only the generation-replay clustering pipeline released
with this paper, better matches the internal structure recovered here and offers a more informative basis for evaluation,
interpretation, and data design. More broadly, the results argue that,
inside a model, what matters is less what a problem is about than what computational approach the model uses to solve it.

{\small
\bibliographystyle{plainnat}
\bibliography{references}
}

\appendix

\section{Text-based clustering baseline}
\label{sec:text-baseline}

A natural question is whether the structure recovered by
generation-replay signatures is already present in the surface text of
the model's generated reasoning. If so, the activation-based features
would add little beyond what a text-only analysis provides. We test
this by clustering TF-IDF/SVD representations of the same generated
reasoning traces using the identical clustering pipeline
($\ell_2$ normalization, 20-dimensional truncated SVD, $\ell_2$
renormalization, K-Means with silhouette-selected
$k\in\{5,\ldots,12\}$).

Table~\ref{tab:text-baseline} compares the two feature spaces across
all 40 model-source cells. Across these cells, both feature spaces
pass C1, confirming that each contains real cluster structure.
However, the structure they capture differs sharply. Text-based
clusters are overwhelmingly topic-aligned: 81.8\% are topic-pure under
the $>$0.70 dominant sub-skill threshold, and the adjusted Rand index
(ARI) against source-provided sub-skill labels is 0.38.
Activation-based clusters show much lower topic alignment: only 36.9\%
are topic-pure, and the ARI is 0.10. Yet, as reported in
Section~\ref{sec:results}, two independent
judges find approach-level coherence in 77--82\% of the
activation-based clusters.

This dissociation confirms that the generation-replay signatures are
not simply recapitulating surface-text similarity. The generated
reasoning text organizes by topic; the internal activations organize by
approach.

\begin{table}[t]
  \centering
  \small
  \begin{tabular}{lrrrr}
    \toprule
    Feature space
      & Clusters
      & Purity $>$0.70
      & Purity $>$0.80
      & ARI \\
    \midrule
    Reasoning text (TF-IDF/SVD)
      & 225 & 81.8\% & 72.4\% & 0.38 \\
    Activations (generation-replay)
      & 225 & 36.9\% & 29.8\% & 0.10 \\
    \bottomrule
  \end{tabular}
  \caption{
  Text-based versus activation-based clustering. Both feature spaces
  use the same pipeline and the same generated reasoning traces across
  the same 40 model-source cells, all passing C1. Purity denotes
  dominant sub-skill purity; ARI is the adjusted Rand index against
  source-provided sub-skill labels. Text clusters are strongly
  topic-aligned; activation clusters are not. Combined with the C2
  result that 77--82\% of activation clusters are
  approach-coherent, this confirms that generation-replay signatures
  capture approach-level structure distinct from surface-text
  similarity.
  }
  \label{tab:text-baseline}
\end{table}

\section{Cluster examples}
\label{sec:cluster-examples}

To make the recovered approach structure concrete, we present two
representative clusters from the C2 evaluation. For each exemplar we
show the abbreviated problem statement and its source-provided
sub-skill label. Full problem texts and generated solutions are
included in the supplementary material.

\paragraph{Cluster A: sequential rate-and-quantity aggregation
(Qwen2.5-7B-Instruct, \texttt{gsm8k}).}
All five exemplars share the source-provided sub-skill label
\texttt{arithmetic\_word}, so topic labels do not distinguish this
cluster from other GSM8K clusters. The C2 evaluation identifies a more
specific shared approach: a forward-chained rate-and-quantity
pipeline in which named sub-quantities are computed in sequence via
multiplication or division (rate $\times$ count, total $\div$ count)
and then combined in a final aggregation step. It also notes that this
schema is not generic to all GSM8K problems, many
of which use comparison, conditional, or equation-solving strategies
instead.

\begin{table}[t]
  \centering
  \small
  \begin{tabular}{p{0.70\linewidth}l}
    \toprule
    Problem (abbreviated) & Sub-skill \\
    \midrule
    Randy drew 5 pictures; Peter drew 3 more; Quincy drew 20
    more than Peter. Total pictures?
    \newline\textit{Approach:} Compute each person's count
    sequentially, sum.
      & arithmetic\_word \\[4pt]
    Bill soaks clothes: 4 min per grass stain and 7 min per
    marinara stain. 3 grass stains, 1 marinara stain. Total time?
    \newline\textit{Approach:} Multiply rate $\times$ count for
    each category, then sum.
      & arithmetic\_word \\[4pt]
    Jett bought a cow for \$600, spent \$20/day on food for 40 days,
    and \$500 on vaccination. Sold it for \$2500. Profit?
    \newline\textit{Approach:} Compute each expense, sum the
    expenses, then subtract from revenue.
      & arithmetic\_word \\[4pt]
    DeShawn made 12 free throws; Kayla made 50\% more;
    Annieka made 4 fewer than Kayla. Annieka's count?
    \newline\textit{Approach:} Compute each person's count
    sequentially from the previous.
      & arithmetic\_word \\[4pt]
    Kim spends 5 min on coffee, 2 min/employee on updates,
    and 3 min/employee on payroll. 9 employees. Total time?
    \newline\textit{Approach:} Compute each task's time,
    then sum.
      & arithmetic\_word \\
    \bottomrule
  \end{tabular}
  \caption{Five nearest-centroid exemplars from a single GSM8K
  cluster. All share the sub-skill label \texttt{arithmetic\_word},
  so topic labels do not differentiate this cluster from others.
  The activation-based clustering groups them by their shared
  reasoning approach: computing named intermediate quantities
  sequentially and then aggregating them.}
  \label{tab:cluster-example-gsm8k}
\end{table}

\paragraph{Cluster B: constraint-to-equation reduction and variable
elimination (DeepSeek-R1-Distill-Qwen-7B,
\texttt{math\_hendrycks}).}
The five nearest-centroid exemplars are drawn from four distinct
sub-skills---precalculus, geometry, algebra, and number
theory---yet all five generated solutions follow the same reasoning
approach. Each solution (i) translates the problem's constraints into
a system of algebraic equations, (ii) eliminates variables by
substitution or by exploiting identities (e.g., Pythagorean identity,
Vieta's formulas, digit-counting symmetry), and (iii) solves the
reduced system to obtain the answer.

\begin{table}[t]
  \centering
  \small
  \begin{tabular}{p{0.62\linewidth}l}
    \toprule
    Problem (abbreviated) & Sub-skill \\
    \midrule
    Given $x + \sin y = 2008$ and $x + 2008\cos y = 2007$ with
    $0 \le y \le \pi/2$, find $x + y$.
    \newline\textit{Approach:} Subtract equations to
    eliminate $x$, substitute into Pythagorean identity,
    factor and solve.
      & precalculus \\[6pt]
    Trapezoid with bases differing by 100; midsegment divides
    area $2{:}3$; find $\lfloor x^2/100 \rfloor$ where
    $x$ bisects the area.
    \newline\textit{Approach:} Express areas algebraically,
    solve ratio equation for base length, derive
    equal-area segment via quadratic relation.
      & geometry \\[6pt]
    $9x^3 + 5ax^2 + 4bx + a = 0$ has three distinct positive
    roots with $\sum \log_2 r_i = 4$; find $a$.
    \newline\textit{Approach:} Apply Vieta's formulas,
    convert logarithm sum to product, substitute into
    product-of-roots relation, solve.
      & algebra \\[6pt]
    Two noncongruent integer-sided isosceles triangles with the same
    perimeter and area; base ratio $8{:}7$; find the minimum
    perimeter.
    \newline\textit{Approach:} Parameterize sides via
    ratio, equate perimeters and areas, substitute into
    Pythagorean height formula, solve quadratic.
      & geometry \\[6pt]
    $S_n$ = sum of reciprocals of nonzero digits from $1$ to
    $10^n$; find smallest $n$ with $S_n \in \mathbb{Z}$.
    \newline\textit{Approach:} Count digit occurrences,
    express $S_n = n \cdot 10^{n-1} \cdot 7129/2520$,
    reduce to divisibility conditions on $n$.
      & number theory \\
    \bottomrule
  \end{tabular}
  \caption{Five nearest-centroid exemplars from a single cluster
  spanning four sub-skills. Despite covering precalculus,
  geometry, algebra, and number theory, all five solutions share
  the same reasoning approach: translate constraints into
  equations, eliminate variables, and solve the reduced system.
  A topic-based clustering would separate these problems; the
  activation-based clustering groups them by shared computational
  approach.}
  \label{tab:cluster-example-hendrycks}
\end{table}

These two clusters illustrate complementary aspects of the main
finding. Cluster~A shows that the activation signatures subdivide a
single topic (GSM8K arithmetic) into approach-specific groups that
topic labels cannot distinguish. Cluster~B shows that the signatures
group problems from different topics together when the model executes
the same approach. In both cases, the generation-replay clustering
captures what computation the model performs, rather than what the
problem is about.

\section{Reproducibility}
\label{sec:reproducibility}

The supplementary material includes the full profiling, clustering,
and analysis code, the raw datasets used for C1/C2 and C3, and the
LLM judge prompts. A README file provides step-by-step instructions
for reproducing all reported results. Reproducing signatures from
scratch requires one forward/backward pass per problem per model on a
single 80\,GB GPU; wall-clock time ranges from approximately two to
fifteen hours per model, depending on model size.

\end{document}